\documentclass[conference]{IEEEtran}
\IEEEoverridecommandlockouts

\usepackage{cite}
\usepackage{amsmath,amssymb,amsfonts}
\usepackage{graphicx}
\usepackage{booktabs}
\usepackage{multirow}
\usepackage{makecell}
\usepackage{siunitx}
\usepackage{xcolor}
\usepackage[hidelinks]{hyperref}
\usepackage{url}
\usepackage{tikz}
\usetikzlibrary{positioning,arrows.meta}

\newcommand{\NF}[1]{#1}

\newcommand{\qh}{\mathbf{q}_{\mathrm{h}}}         
\newcommand{\qr}{\mathbf{q}_{\mathrm{r}}}         
\newcommand{\Mmap}{M}                             
\newcommand{\dvec}{\theta}                        
\newcommand{\rrule}{\rho}                         
\newcommand{\ik}{\mathrm{IK}_{\rrule}}            
\newcommand{\Kh}{K^{\mathrm{h}}}                  
\newcommand{\Kr}{K^{\mathrm{r}}}                  

\begin{document}

\title{PACT: Posture-Aligned Co-Design Technique for \\ Hand Configuration and Retargeting in \\ Multi-Finger Imitation Learning}

\author{\IEEEauthorblockN{Koyo Sasazaki, Satoshi Funabashi, Pranav Ponnivalavan,\\
Alexander Schmitz, Tetsuya Ogata and Shigeki Sugano}
\IEEEauthorblockA{Waseda University}%
\thanks{This work was supported by JST Moonshot R\&D Grant Number \mbox{JPMJMS263C} and by Grant-in-Aid for Scientific Research (B) (KAKENHI) from Japan Society for the Promotion of Science (JSPS) with a grant number of \mbox{24K03003}. All authors are with Waseda University, \mbox{Okubo 3-4-1}, Shinjuku, Tokyo \mbox{169-8555}, Japan (\mbox{koyossthekey@fuji.waseda.jp}).}%
}

\maketitle

\begin{abstract}
The fidelity of teleoperated demonstrations used for imitation learning is determined by how faithfully a robot hand can replicate the operator's posture. The posture retargeting law that the robot will ultimately execute is typically selected only once the hardware has been finalized. Conventional hand designs are derived from established mechanical principles, so they often neglect human-like grasping postures, which constrains the potential of learning-based dexterous manipulation. We introduce PACT, an optimization based on human grasping postures that jointly optimizes the mechanical parameters of a 17-servo robot hand (link lengths, joint offsets, mounting angles, and base placements) and its nonlinear posture retargeter (a linear human-to-robot joint map, per-joint posture-IK weights, and damping). The optimization enforces full-finger correspondence (MCP, PIP, DIP, and tip of each finger) to human poses from the DexYCB dataset, and is also constrained by physical requirements such as servo interference, assembly clearances, and flat-hand protrusion, so the resulting design can be physically assembled. We fabricate the PACT Hand and validate it through teleoperation experiments and an imitation-learning demonstration. The proposed hand reproduces all human hand keypoints with a kinematic root-mean-square error (RMSE) of 3.9 mm, compared to 12.8 mm for the LEAP Hand baseline, and keeps all fingertips within 10 mm on every grasping posture, while the baseline achieves this for only 21.6\% of the poses. Under matched per-joint torque limits, the proposed hand lifts twice the baseline payload, 2 kg against 1 kg, and deforms a held paper cup seven times less, reaching a deformation level comparable to a human hand. It also achieves an 86\% grasping success rate on YCB daily objects, compared to 64\% for the baseline. Finally, as a demonstration, an imitation-learning policy trained on demonstrations collected with the PACT Hand achieved a 75\% success rate at held-out placements in a can pick-and-place task. Results visualizations and videos at \url{https://pact-hand-2027.github.io}
\end{abstract}

\section{Introduction}
\label{sec:intro}

\begin{figure}[t]
  \centering
  \NF{\includegraphics[width=1.0\linewidth]{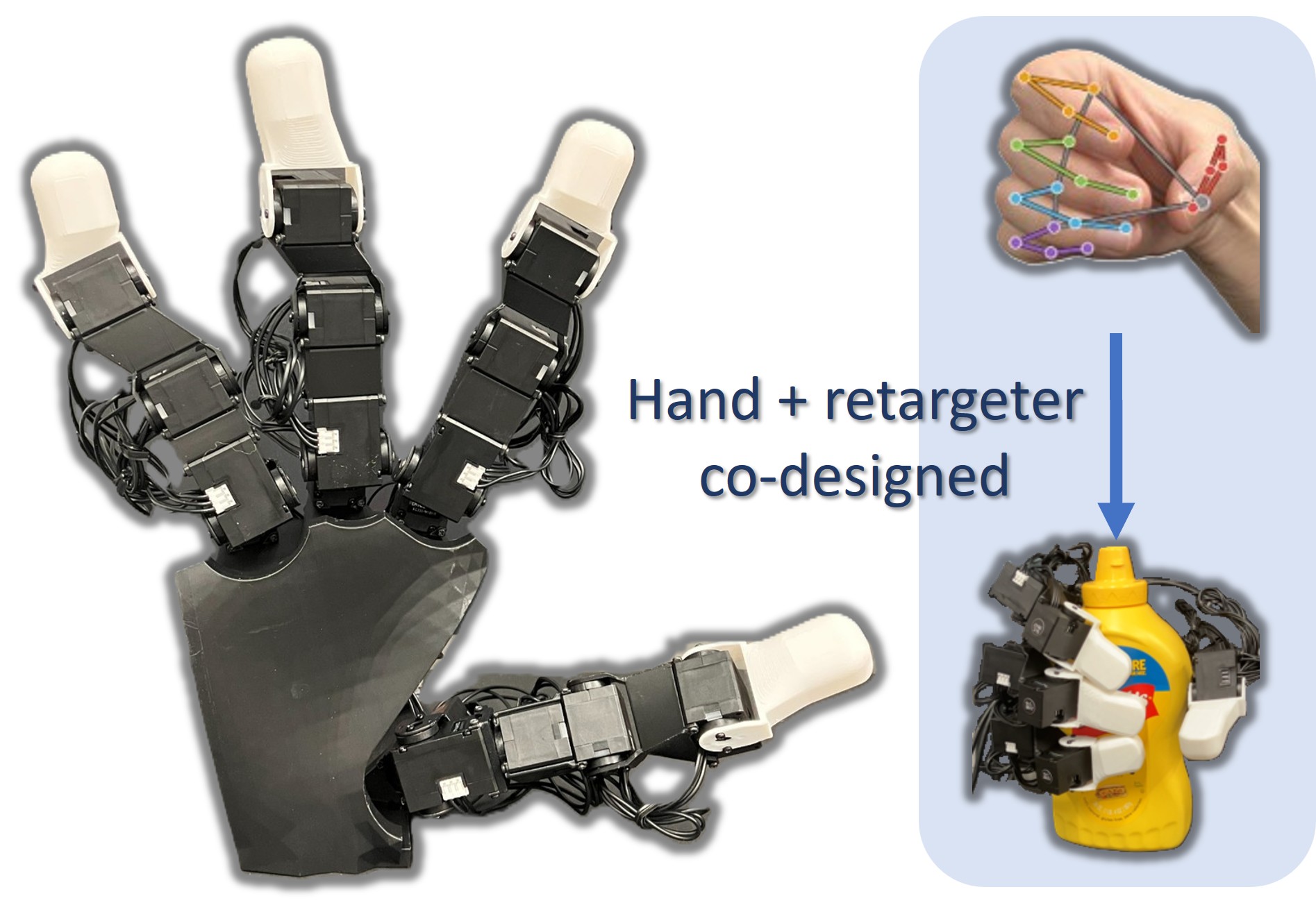}}
  \caption{The PACT Hand (left), a 17-servo anthropomorphic hand co-designed with its posture-IK retargeter. At runtime (right), the same retargeting law that served as the design objective maps a bare-hand posture from a single RGB camera onto the hand, shown grasping a mustard bottle.}
  \label{fig:teaser}
\end{figure}

The quality of teleoperated demonstrations upper-bounds the performance of dexterous manipulation learning, and is itself limited by how faithfully the robot hand reproduces human postures. Kinematic mismatches between humans and robots are typically resolved in software by a retargeting solver that maps observed human hand poses onto the available robotic hardware~\cite{handa2020dexpilot,qin2023anyteleop}. When the hardware cannot follow the human's posture, because its joint configuration and actuator sizes differ substantially from the human hand, the solver cannot intrinsically close that gap.

Recent work has therefore begun to shape the hardware itself around human motion. Hand morphologies have been optimized to reproduce recorded human fingertip trajectories~\cite{yi2026generating}, and robot hand body-to-controller co-design has been achieved, including on physical hands~\cite{chen2021hardware}. Yet in all of these, the retargeting algorithm that actually runs during teleoperation is not considered in the co-design of robot hands and their controllers. The gap this leaves is practical. An optimized hand cannot minimize the error against recorded human postures unless its retargeter is optimized with it, and human-like manipulation needs contact along the phalanges and the palm, not at the fingertips alone.

We formulate a hand design framework in which the design objective is the residual of the posture-IK law that later runs on the robot for teleoperation. Specifically, a linear human-to-robot joint map, the per-keypoint weights of the IK objective, and its damping are optimized jointly with hand hardware parameters such as link lengths, joint offsets, mounting angles, and base placements. Matching only fingertips leaves the intermediate joints free to drift into non-human configurations.

The correspondence is enforced at the level of the full finger grasping posture. Each keypoint of every finger, meaning its MCP, PIP, DIP, and tip, is optimized to spatially align with the corresponding human keypoint across a distribution of real grasping postures~\cite{chao2021dexycb}.

Since the buildability of an actual robot hand is essential, interference between real servo housings, a measured assembly clearance, and flat-hand palmar protrusion are incorporated into the same optimization objective. We 3D-print and assemble the 17-servo PACT Hand, the hardware realization of our proposed \emph{P}osture-\emph{A}ligned \emph{C}o-design \emph{T}echnique. We deploy the same retargeter to control it during teleoperation and imitation learning.
Our contributions to designing an optimized robot hand for imitation learning are:
\begin{itemize}
  \item a co-design formulation in which the deployed nonlinear posture retargeter and the hand hardware configuration are jointly treated as design variables, with the residual of that same deployed retargeter serving as the design objective, trained over a real human grasp-pose distribution.
  \item full-finger posture correspondence (MCP/PIP/DIP/tip) as the matching criterion, rather than fingertip positions alone.
  \item a staged optimization scheme that integrates servo interference, assembly clearance, and flat-hand protrusion into a single design framework without compromising tracking accuracy, validated by physically building the hand.
\end{itemize}

\section{Related Work}
\label{sec:related}

\subsection{How Anthropomorphic Hands Are Designed}
Servo-driven, printable hands have made dexterous manipulation research broadly reproducible~\cite{shaw2023leap,shaw2025leapv2,christoph2025orca}, and several recent designs are built for teleoperated data collection. Their kinematic parameters are typically chosen from anthropometric and clinical hand data~\cite{zhu2026midas}, copied directly from a human model~\cite{junge2025adaptteleop}, or fixed by mechanism-level engineering judgement. Human similarity is then assessed afterwards as a workspace-overlap score~\cite{liarokapis2013quantifying} or a range-of-motion comparison~\cite{junge2025adaptteleop}. 

Human posture statistics have long directly informed hardware design. Postural synergies, the low-dimensional subspace extracted from recorded hand postures by principal component analysis, are widely used to command many joints from few coefficients, and have also been embodied in hardware as coupled transmissions and spring stiffnesses~\cite{brown2007interfinger,catalano2014adaptive,xu2014posturalsynergies}. However, these approaches still require manually defining which human quantity a given joint should track.

\subsection{Human-Data-Driven Hand Design}
To decide that correspondence automatically, a separate line of work uses recorded human motion directly as the design target. Yi \emph{et al.}~\cite{yi2026generating} state a principle we also adopt, that a design should be optimized under the controller that will run after fabrication, and optimize linkage geometry together with a joint trajectory so that the mechanism reproduces recorded human motions. A simple robot IK resolved for each new joint input at each timestep exposes no parameters to the design problem and imposes no correspondence on the joints between the finger base and the tip. Building on this idea, motion-conditioned design has been generalized across different embodiments~\cite{ha2026transformer}, though for fixed-base arms, quadruped manipulators, and wheeled bimanual manipulators rather than for robot hands. Other work has searched soft-hand parameters for grasp success with a gradient-free method, guided by a reward model learned from simulated grasps that replay teleoperation data~\cite{bai2026rewardmodels}.

Fabrication constraints should also be included in optimizations for real robot assembly. Passive grippers have been optimized with collision-free insertion enforced during the search, then 3D-printed and tested with object picking~\cite{kodnongbua2022passive}. Mechanism dimensions and motor placement have been co-optimized for buildable devices~\cite{ha2018computational}. Dexterous-hand parameters have also been optimized using a parameterization whose output geometry is directly fabricable~\cite{mirzaee2026handcdo}. Yet none of these validates a fabricable robot hand for imitation learning, which is critical since learned robot motions depend on the controller as well as on the hardware. One of the closest studies jointly optimizes a fingertip cover geometry with the corresponding control, and applies the result to imitation learning on real hardware~\cite{ye2025powerprecision}. Its design objective is grasp quality, however, rather than correspondence to the human operator's posture, so the deployed retargeting map remains outside the optimization.

\subsection{Co-Design of Morphology and Control}
Co-optimizing hardware and control has been investigated across various robot embodiments. For hands, various elements have been co-designed. These include soft-hand morphology with open-loop commands~\cite{deimel2017automated}, finger geometry with grasp-type torque programs~\cite{pan2021emergent}, and palm and finger shape with an RL expert~\cite{mannam2024design}. Further examples are tendon-driven chains with controllers~\cite{islam2024tentamorph}, finger link lengths and placement with RL policies through a differentiable hardware proxy~\cite{he2024morph}, and graph-structured hand morphologies with morphology-conditioned policies at scale~\cite{fay2026dextra}. One such method co-optimizes the transmission and spring parameters of an underactuated hand with a neural control policy, and deploys both on real hardware~\cite{chen2021hardware}.

Although co-design of robot hardware and its controllers has been investigated, none of these formulations contains a human-to-robot correspondence. The robot is never required to reproduce a specific operator's posture, i.e., retargeting, at run time for imitation-learning data collection. This is important because a retargeter can affect the controllability of a robot hand regardless of its optimal hardware design for dexterous manipulation.

\subsection{Retargeting and the Quality of Teleoperated Data}
Retargeting algorithms themselves have been extensively optimized. Vision-based retargeting solves a per-frame correspondence onto a fixed hand~\cite{qi2026genhand}. ByteDexter's palm is non-anthropomorphic because of actuator packaging, and its runtime retargeting formulation compensates for the discrepancy~\cite{wen2025bytedexter}. RAPID Hand deploys a retargeting objective that aligns every phalanx alongside its hand and perception stack, but the retargeter's parameters are not design variables of the mechanism~\cite{wan2025rapid}. 

In parallel, demonstration quality has been improved through dedicated hardware. This includes wearable interfaces~\cite{dexumi2025}, high-fidelity collection rigs~\cite{wei2026hifiumi}, and distributed tactile feedback that measurably tightens the distribution of teleoperated trajectories~\cite{kota2026missing}. These results support the premise that hardware choices measurably affect demonstration quality. However, to the best of our knowledge, no prior hand-design method jointly optimizes mechanism parameters with retargeting-map or posture-IK parameters under full-finger human-pose correspondence for improving teleoperation and imitation-learning quality.

\section{Method}
\label{sec:method}

\begin{figure*}[t]
  \centering
  \NF{\includegraphics[width=\textwidth]{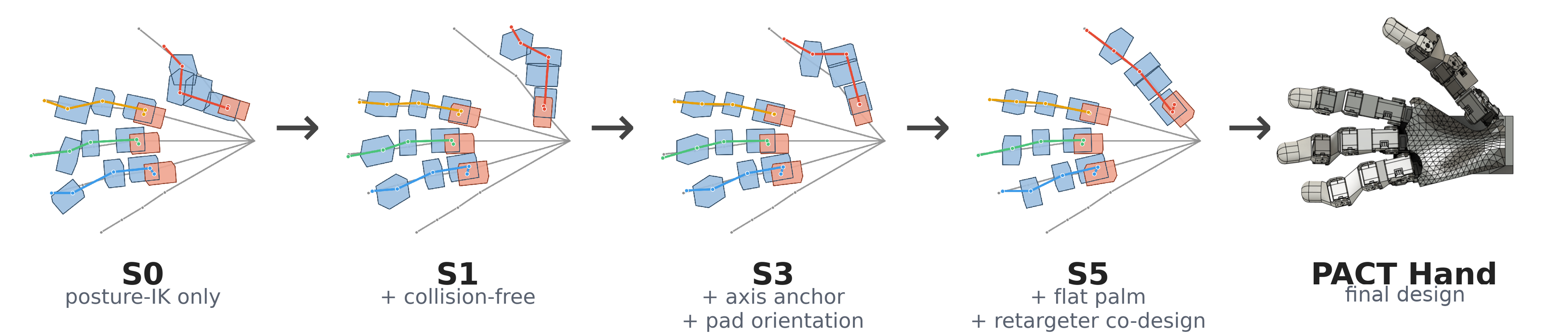}}
  \caption{Fabrication constraints reshaping the hand, stage by stage. Optimized joint chains and as-built servo volumes are overlaid on the human skeleton at the flat-hand pose after stages S0, S1, S3, and S5 (blue: flexion, salmon: abduction; gray: human skeleton at $1.56\times$; annotations list constraints added since the previous panel). Right: the final design.}
  \label{fig:pipeline}
\end{figure*}

This section formulates the co-design problem, defines the deployed retargeting rule, and describes the staged introduction of fabrication constraints that renders the design manufacturable. Fig.~\ref{fig:pipeline} shows the design converging to the human skeleton as the constraints are added.

\subsection{Co-design Problem}
\label{sec:method-problem}

Let $p$ be a human hand pose drawn from a grasp-pose distribution $\mathcal{D}$, represented as MANO joint angles~\cite{mano2017}. It is mapped into the design frame at a scale of 1.56, the mean ratio of the LEAP Hand's total kinematic chain length, housings and fingertip block included and taken from its official URDF~\cite{shaw2023leap}, to the corresponding human finger lengths. The mechanism is parameterized by $\dvec$, which comprises per-joint link lengths, twists, joint and axis offsets, and each finger base's pose on the palm. The retargeting rule is parameterized by $\rrule = (\Mmap, w, \lambda^2)$, defined below. With $\Kh_k(p)$ the $k$-th human keypoint and $\Kr_k(\qr;\dvec)$ the corresponding robot keypoint under forward kinematics, the co-design problem is

\begin{multline}
\min_{\dvec,\,\rrule}\;
\mathbb{E}_{p\sim\mathcal{D}}\!\left[
\sum_{k}\bigl\lVert
\Kr_k\!\bigl(\ik(p;\dvec);\dvec\bigr)-\Kh_k(p)
\bigr\rVert^2\right]\\
+\;\sum_{j}\gamma_j\,C_j(\dvec),
\label{eq:codesign}
\end{multline}

where $\ik(p;\dvec)$ is the deployed posture IK defined in Sec.~\ref{sec:method-rule}, $k$ ranges over the MCP, PIP, DIP, and tip of every finger, and the $C_j$ are the physical fabrication constraints introduced one at a time in Sec.~\ref{sec:method-ladder}. The expectation is taken over recorded human grasping poses from DexYCB (described in Sec.~\ref{sec:method-data}), not over a geometric workspace, so the hand is optimized for postures humans actually use. The IK solver inside \eqref{eq:codesign} is the same implementation, run for the same number of iterations, that later controls the physical hand.

\subsection{Deployed Retargeter}
\label{sec:method-rule}

The deployed retargeter maps a human pose to robot joint angles in two steps. First, a linear joint map gives a seed $\qr^{0} = \Mmap\,\qh(p)$, where $\Mmap$ performs sign calibration and establishes a coarse correspondence. Second, a damped-least-squares posture IK refines this seed over 20 iterations, minimizing the weighted keypoint residual $\sum_k w_{c(k)}\lVert \Kr_k(\qr)-\Kh_k(p)\rVert^2$ with damping $\lambda^2$. Here $c(k)$ maps each keypoint to a shared weight per keypoint type, with one weight each for the PIP, DIP, and tip keypoints. The MCP keypoint is given zero weight inside the IK, since its position is determined by where the finger base is mounted on the palm, not by the joint angles. Its correspondence is instead added to \eqref{eq:codesign} as a separate term, stage S2 in Sec.~\ref{sec:method-ladder}.

In the final design stage, $\Mmap$, the weight classes $w$, and $\lambda^2$ become design variables optimized jointly with $\dvec$. The weights affect only the solver's step direction. The residual in \eqref{eq:codesign} is always measured with fixed unit weights, so the optimizer cannot reduce the reported error simply by shrinking $w$. The optimized retargeter is saved with the hardware design and loaded unmodified at runtime. Design, evaluation, and physical operation all execute the same $(\Mmap, w, \lambda^2)$.

\subsection{Staged Introduction of Fabrication Constraints}
\label{sec:method-ladder}

Problem \eqref{eq:codesign} must produce a hand that can be physically assembled. Empirically, we impose the fabrication requirements not simultaneously but sequentially, as a sequence of optimization stages S0 through S6. Each stage adds one requirement as a penalty $C_j$ to \eqref{eq:codesign} and starts from the previous stage's solution (a warm start). Imposing every constraint simultaneously from scratch frequently failed to converge. The staged form instead absorbs one requirement at a time, while tracking accuracy is monitored.

The stages are as follows. \textbf{S0} uses tracking accuracy alone, establishing the accuracy ceiling of the kinematic family. \textbf{S1} adds a requirement of no collision between any two servo housings, enforced across the whole pose distribution. \textbf{S2} requires the robot MCP axes to coincide with the human MCP centers. \textbf{S3} requires fingertip pads to stay within $20^\circ$ of the human pad orientation. \textbf{S4} forbids any geometry from protruding below the flat palm. \textbf{S5} adds the retargeter parameters $\rrule$ to $\dvec$ as design variables. \textbf{S6} adds a measured assembly clearance between stacked servos.

From S1 onward, every stage also carries an accuracy guard. This is a one-sided penalty on any per-finger tracking term that exceeds its value at the start of the stage by more than 10\%. A stage therefore cannot satisfy its new constraint by sacrificing the tracking accuracy already achieved. Collision avoidance at S1 pushes the thumb away from the human skeleton in Fig.~\ref{fig:pipeline}, and the MCP axis anchor added at S2 pulls it back by S5.

\subsection{Implementation}
\label{sec:method-impl}

The index, middle, and ring fingers share a two-axis MCP module, and the thumb a three-axis CMC module. Distal joints are serial servo chains, mounted dorsally so the housings rest opposite the finger pads, with the final pair of servos stacked case to case. The whole pipeline, including forward kinematics, the unrolled 20-iteration IK, and all constraint terms, is differentiable in PyTorch at float64. Each stage runs 32 random restarts of 400 optimizer iterations.

\subsection{Dataset}
\label{sec:method-data}

Human poses come from 234 DexYCB grasping sequences~\cite{chao2021dexycb} spanning all 10 subjects and 8 YCB objects. Frames are sampled at stride 3, and MANO fits with a residual above \SI{8}{mm} are discarded, leaving 5,117 poses. Every pose is represented by joint angles on a common mean MANO hand shape, so the distribution varies in grasping posture, not in subject hand size. The 234 sequences were not selected by hand side, fit quality, or object identity. Object identity never enters \eqref{eq:codesign}, and Sec.~\ref{sec:exp-accuracy} evaluates the design on twelve YCB objects unseen during design.
Stages S0 through S6 train on 300 poses selected by farthest-point sampling from the 5,117-pose corpus to keep the optimization fast. The final design is then re-optimized, warm-started, on a 2,000-pose subset, which a kinematic-feasibility filter reduces to the 1,788 training poses used for evaluation. The evaluation poses kept separate from training are described in Sec.~\ref{sec:exp-accuracy}.

\section{Hand and Teleoperation System}
\label{sec:hand}

The optimized mechanism is fabricated as the physical PACT Hand: four digits, an index, middle, and ring finger with four actuators each, plus a thumb with five actuators, built from Dynamixel servos and 3D-printed brackets. The four-digit layout follows the LEAP Hand baseline, which likewise omits the little finger, so the two hands are compared at the same digit count and a closely matched actuator budget (17 for PACT vs.\ 16 for LEAP). All bracket geometry is exported directly from the optimized parameters without manual re-dimensioning. The hand matches the LEAP Hand in total kinematic chain length per finger (153--161\,mm for PACT vs.\ 145--153\,mm for LEAP) but distributes it with human anatomical proportions, whereas LEAP concentrates it in actuator housings and a rigid \SI{50}{mm} fingertip block. PACT's fingers therefore appear longer at the same chain-length scale.

The design problem also exhibits scale equivariance. Link lengths, keypoints, and housing dimensions share a common factor, whereas twists, offsets, pad angles, and joint limits do not. The 1.56 scale was chosen to match the LEAP baseline, not as a lower limit. A smaller hand may be feasible even with the same servos, at some cost in tracking accuracy, since the binding constraints are housing interferences. Conversely, the largest servo housing that admits a human-sized hand at a target accuracy can be computed by re-running the staged optimization at smaller scales. The housing dimensions in the optimization were set to those of the Dynamixel XC330-M288-T, and the servos of the physical hand share this housing, so the optimized geometry carries over unchanged. The housing model is a configurable parameter of the design pipeline.

Operator poses for teleoperation are estimated from a single RGB camera. WiLoR~\cite{wilor2025} regresses MANO parameters, the posture-IK rule $(\Mmap, w, \lambda^2)$ saved with the design converts them to joint targets, and the targets stream to the servos at roughly \SI{23}{Hz}. No per-user calibration or parameter tuning was performed in any experiment. The retargeter saved with the design was deployed as-is. For the comparisons in Sec.~\ref{sec:exp-teleop}, all hands run current-based position control with the commanded current set from each motor's torque constant, so every joint of every hand operates under an identical torque cap. We verified the caps post hoc with a no-load speed sweep of the baseline's servos.

\section{Experiments}
\label{sec:experiments}

\subsection{Retargeting Accuracy}
\label{sec:exp-accuracy}

\looseness=-1 Accuracy is evaluated using forward kinematics on the commanded joint configurations, rather than on measured servo-tracking errors. Specifically, each human pose is retargeted to the corresponding robot hand, the resulting robot keypoints are computed, and the residuals are reported in millimeters at human scale, with errors measured in the design frame divided by 1.56. All hands are evaluated against the same human keypoints scaled by a factor of 1.56, using the same normalization procedure. We also report two summary statistics. HPCR($\varepsilon$) is the Human Pose Coverage Rate, the fraction of poses whose worst-finger fingertip error is within $\varepsilon$, evaluated at $\varepsilon = \SI{10}{mm}$. The MCP/PIP and IP/DIP columns report per-keypoint-class RMSE, pooled over all fingers and poses on the training set.

The \emph{LEAP Hand} baseline runs the identical posture-IK pipeline on the published LEAP kinematics~\cite{shaw2023leap}, its fingertip frame offsets verified against the physical hardware. The \emph{tip-only} baseline is optimized with our full pipeline and all physical constraints, but its human-matching objective is limited to fingertip keypoints, at design time and in the runtime IK. It features four thumb actuators, 16 total, matching the LEAP budget, instead of five. This is an outcome of the methodology rather than a handicap. The fifth actuator controls distal pad orientation, which the tip-only objective ignores, and the 16-actuator design already saturates fingertip positioning accuracy.

The training set consists of the 1,788 poses on which the final design was re-optimized, obtained by applying the kinematic-feasibility filter to the 2,000-pose farthest-point-sampled subset of Sec.~\ref{sec:method-data}. The test set contains 277 poses selected from the original 5,117-pose corpus and shares no pose with the training set. All frames within $3^\circ$, all-joint $L_\infty$, of any training pose were first removed, which excluded 2,275 near-duplicate frames originating from neighboring video frames. The test set therefore represents data drawn from the same distribution as the training data, rather than data involving unseen subjects or objects. Each hand is evaluated using the retargeting rule associated with its design. Our method uses the jointly designed rule, whereas each baseline uses the IK weights employed during its design, LEAP $[0,1,1,1]$ and tip-only $[0,0,0,1]$.

\begin{table}[t]
\caption{Retargeting accuracy, in human-scale mm, on the training and test sets.}
\label{tab:accuracy}
\centering\footnotesize\renewcommand{\arraystretch}{0.93}
\setlength{\tabcolsep}{2.2pt}
\NF{\begin{tabular}{llrrrrrr}
\toprule
Hand & Set & $n$ & \makecell[r]{All-kp\\RMSE} & \makecell[r]{Tip\\RMSE}
 & \makecell[r]{MCP/\\PIP} & \makecell[r]{IP/\\DIP} & HPCR \\
\midrule
PACT (17)     & train    & 1788 & \textbf{3.90} & 1.84 & \textbf{4.39} & \textbf{4.80} & \textbf{100.0\%} \\
PACT (17)     & test &  277 & \textbf{3.88} & 1.66 & --- & --- & \textbf{100.0\%} \\
LEAP Hand     & train    & 1788 & 12.82 & 13.45 & 14.13 & 10.60 & 21.6\% \\
LEAP Hand     & test &  277 & 13.09 & 13.59 & --- & --- & 7.9\% \\
Tip-only (16) & train    & 1788 & 23.66 & \textbf{2.46} & 26.33 & 31.31 & 99.5\% \\
Tip-only (16) & test &  277 & 22.79 & $<$0.01 & --- & --- & 100.0\% \\
\bottomrule
\end{tabular}}
\end{table}

\looseness=-1 Table~\ref{tab:accuracy} summarizes the results. The co-designed hand tracks the full finger posture at an all-keypoint RMSE of \SI{3.90}{mm}. It reproduces every pose with every fingertip within \SI{10}{mm}, reaching 100\% HPCR, and maintains this performance on the test set at \SI{3.88}{mm}. The LEAP baseline yields an all-keypoint RMSE of \SI{12.82}{mm} and covers only 21.6\% of poses with the same solver and pose set. Beyond the aggregates in Table~\ref{tab:accuracy}, a pose-wise analysis gives best, median, and worst all-keypoint RMSEs of 3.4, 3.9, and \SI{4.9}{mm} for PACT, against 6.9, 12.0, and \SI{35.9}{mm} for LEAP Hand, so the PACT Hand's worst training pose is still more accurate than the baseline's best. The tip-only baseline shows the opposite pattern. With four DOF per finger and a single 3D target, its fingertips are essentially exact, with test-set tip RMSE below \SI{0.01}{mm}, while its intermediate joints drift to \SI{23.66}{mm}. The MCP/PIP and IP/DIP columns of Table~\ref{tab:accuracy} show fingertip saturation achieved at the cost of intermediate-joint drift, whereas posture matching holds every keypoint class below \SI{5}{mm} at a fingertip cost of at most \SI{1.7}{mm}.

We also evaluated all three hands on 1,421 feasible grasps of the twelve YCB objects absent from the design set. Every pose is at least $3^\circ$, all-joint $L_\infty$, from every training pose, with a median of $12^\circ$. PACT degrades by \SI{0.16}{mm} in all-keypoint RMSE, from 3.90 to \SI{4.06}{mm}, and reaches 99.6\% HPCR, with per-object all-keypoint RMSEs within 3.89 to \SI{4.19}{mm} for eleven of the twelve objects. The exception is the pitcher, at \SI{4.74}{mm}, whose handle grasp has no counterpart among the design objects. The tip-only ablation degrades three times as much in MCP/PIP RMSE, from 26.33 to \SI{29.33}{mm}, at unchanged fingertip error. Fingertip accuracy alone does not transfer to joint-level similarity to the human hand. The LEAP Hand's all-keypoint RMSE on these unseen objects, \SI{11.16}{mm}, is in fact lower than its \SI{12.82}{mm} on the eight design objects, because the unseen objects elicit deeper grasps that favor its shorter phalanges. This drop does not indicate better generalization by the LEAP Hand. The two object sets elicit different distributions of grasp posture, so hands should be compared on the same object set.

Resetting $\rrule=(\Mmap,w,\lambda^2)$ to its initialization, $\Mmap_0$ with default weights $[0,1,1,1]$, on the finalized hardware isolates the contribution of the tuned retargeter from that of the hardware. Even under this default retargeter, the PACT Hand achieves an all-keypoint RMSE of \SI{3.47}{mm} and 100\% HPCR on both sets, still $3.7\times$ more accurate than LEAP Hand.
What the co-designed retargeter contributes is accuracy at the contact interface. It reduces tip RMSE from 2.88 to \SI{1.84}{mm}, and from 2.73 to \SI{1.66}{mm} on the test set, for a \SI{0.43}{mm} all-keypoint concession. Shifting accuracy toward the contact points therefore costs \SI{0.4}{mm} when generated by the retargeter, and nearly \SI{20}{mm} when the hardware geometry itself is committed to the fingertips, as in the tip-only design.

\subsection{Teleoperated Grasping}
\label{sec:exp-teleop}

\looseness=-1 
In this experiment, all hardware comparisons run under the torque-matched protocol of Sec.~\ref{sec:hand}. The hypothesis is that at an identical per-joint torque cap, posture correspondence enables the PACT Hand to establish contact over the finger surfaces, the pads as well as the tips, and thus to hold heavier and more fragile objects. The hands run the same retargeting pipeline, each under the IK weights it was optimized with. Each hand was mounted on a UR5e executing one fixed lift of \SI{150}{mm} at \SI{50}{mm/s} with a \SI{3}{s} hold. The tip-only design was also built and used for the experiment.

\begin{figure*}[t]
  \centering
  \NF{\includegraphics[width=0.33\textwidth]{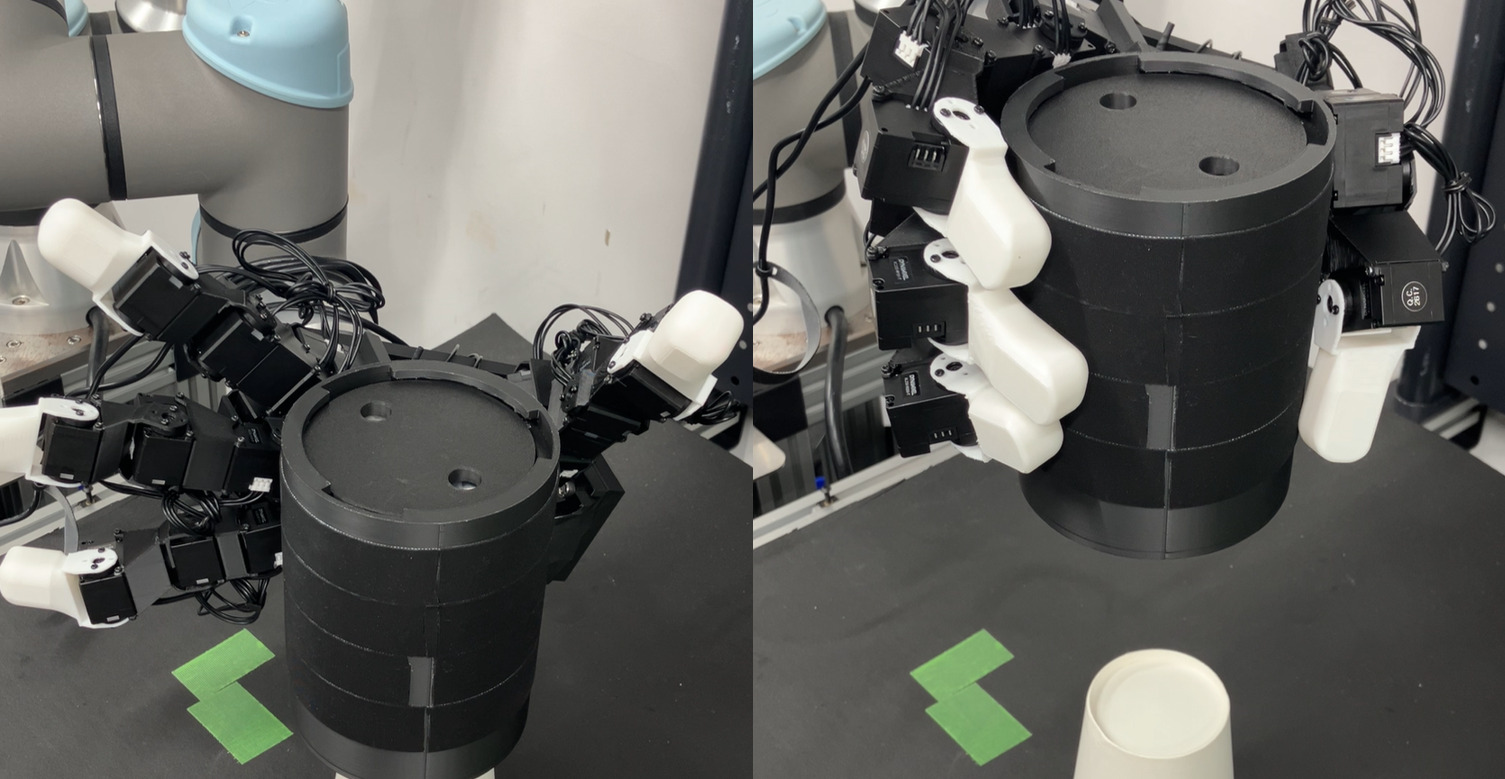}}\hfill
  \NF{\includegraphics[width=0.63\textwidth]{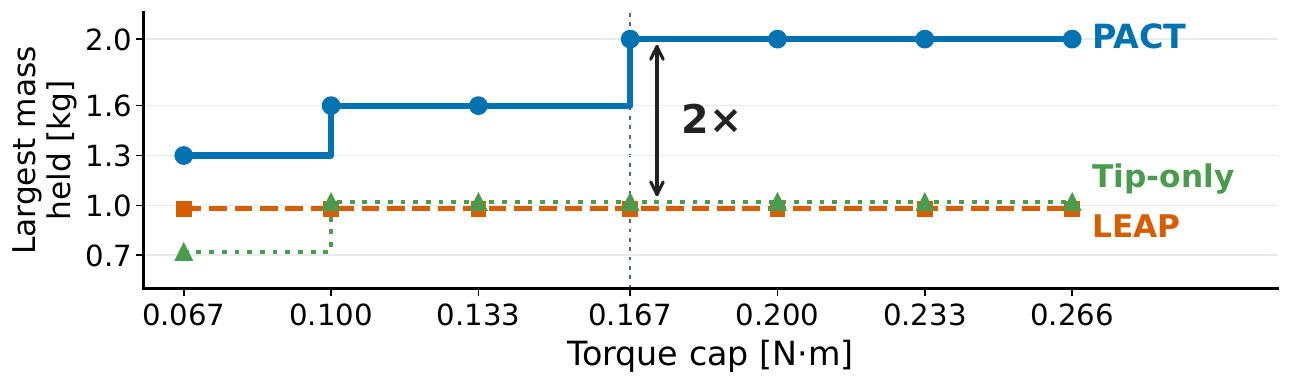}}
  \caption{Payload under a per-joint torque cap. Left: the
  replayed grasp and lift of the instrumented cylinder.
    Right: the largest mass held reliably at each cap for each robot hand.
  The two baseline curves are offset by \SI{20}{g} so that they remain
  distinguishable. The LEAP Hand's occasional \SI{1.3}{kg} holds, at most
  three of five trials, are not counted as reliable.
  }
  \label{fig:payload}
\end{figure*}

The first experiment measures how much mass each hand can lift and hold under a torque cap. The object is a rigid 3D-printed replica of the YCB master chef can, \SI{102}{mm} in diameter, and loaded so the center of mass sits at the grasp band equipped to each hand. Six masses (400--\SI{2000}{g}) $\times$ seven caps (0.067--\SI{0.266}{N.m}) give a $6 \times 7$ grid for all three hands. To isolate the torque cap as the only variable, the grasp was teleoperated once per hand and the recorded command replayed identically in every trial. 
At least five trials were conducted for each combination of mass and torque cap.
A trial succeeds if the cylinder is held at the end of the lift. Fig.~\ref{fig:payload} shows the replayed lift and the resulting capacity curves.
For the PACT Hand, masses from 400 to \SI{1300}{g} succeed at the lowest cap tested, \SI{1600}{g} needs \SI{0.100}{N.m}, and \SI{2000}{g} needs \SI{0.167}{N.m}.
LEAP Hand instead saturates near \SI{1}{kg} regardless of the cap, and the tip-only hand stops at the same place, \SI{1000}{g} at \SI{0.100}{N.m}, with nothing heavier succeeding under any cap. At the common \SI{0.167}{N.m} cap, the capacity is \SI{2000}{g} for PACT Hand, against \SI{1000}{g} for both baselines. At \SI{1300}{g} and \SI{0.167}{N.m}, PACT succeeds 5/5, while LEAP manages 0/6 and the tip-only hand 0/5.

How the tip-only hand fails is informative, even though its overall capacity matches that of the LEAP Hand. Of its 31 failures at \SI{1300}{g}, twenty are recorded as lifted and then dropped, and only one as a failure to lift. This implies that the hand shape, especially phalanges which produce contact areas with a grasped object, was not optimized enough for holding a heavy object for several seconds.

\looseness=-1 In the second experiment, the three hands are evaluated on whether a compliant object can be lifted and held without excessive deformation. This is to confirm the hypothesis that the hand is optimized based on human hand grasping postures and can achieve conformity to a grasped object shape which leads to broader contact areas and less grasping force to cause deformation. A paper cup loaded to \SI{400}{g} is grasped.
The grasp is replayed identically in every trial, fixing the operator's finger closing distance. The cup rim is cut off to standardize the cup, and a red band on the inner wall is put as a circle detector in the ellipse-fitting of the image-processing method to compare the deformation for each hand.
We report the circularity $4\pi A/P^{2}$ of the band after the affine map that takes the ellipse fitted to the \emph{same trial's} pre-grasp image to the unit circle. A value of 1.000 therefore corresponds to that trial's own undeformed rim.

\begin{figure}[t]
  \centering
  \NF{\includegraphics[width=1.0\linewidth]{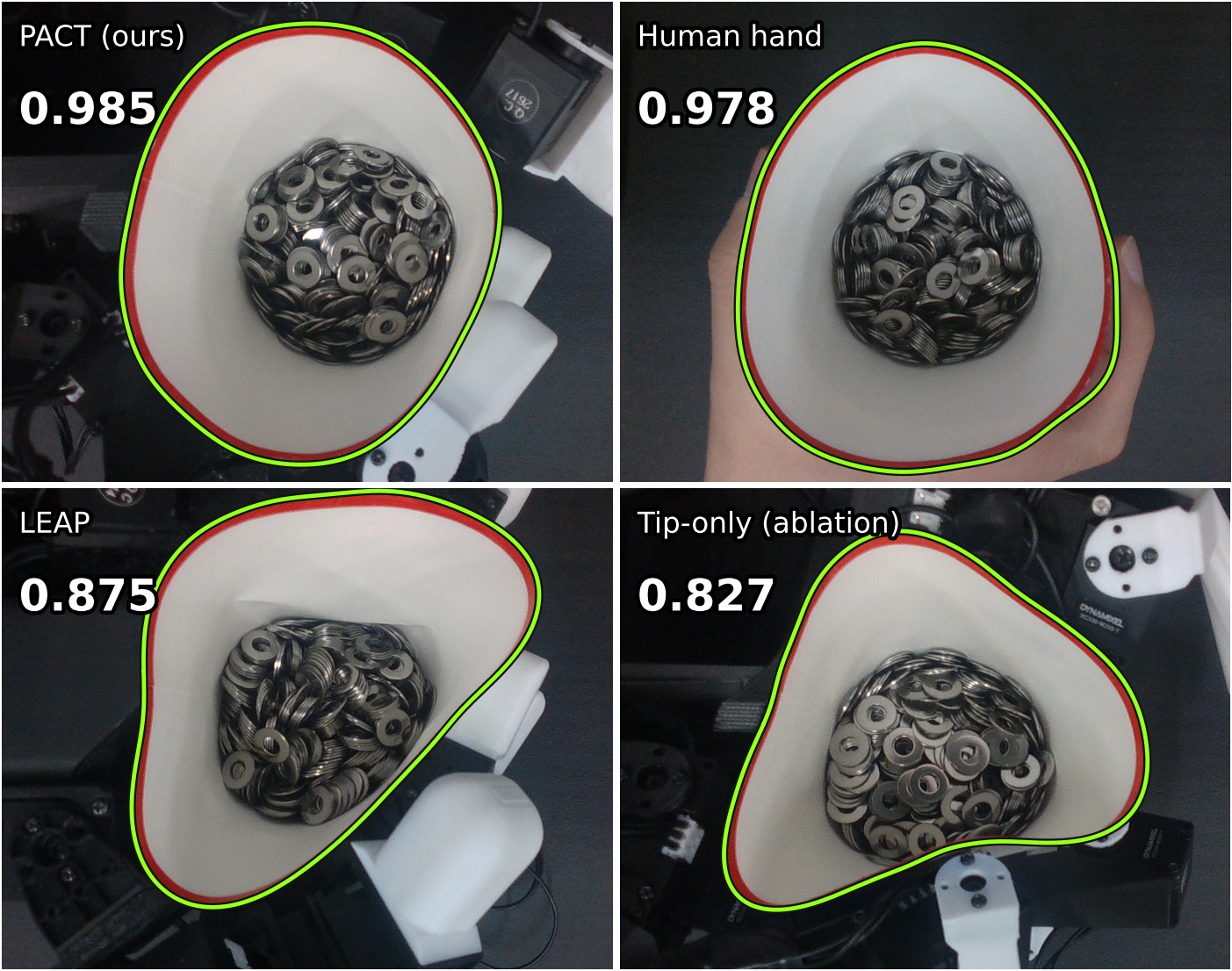}}
  \caption{Rim deformation holding the \SI{400}{g} cup, trials at
  \SI{0.167}{N.m}. The detected band is shown in green, with
  pre-frame normalized circularity at top left.}
  \label{fig:cup}
\end{figure}

Five caps (0.133--\SI{0.266}{N.m}) $\times$ five trials give 25 trials per hand, and five grasps were also recorded with a human hand for reference. All 75 robot trials held the cup, so what separates the hands is deformation, not holding. LEAP Hand flattens the rim to $0.8752 \pm 0.0120$, while PACT Hand stays at $0.9847 \pm 0.0020$ (Fig.~\ref{fig:cup}). This is a $7\times$ larger departure from a circle, 0.125 against 0.015. The PACT Hand is on par with the human hand, at $0.9777 \pm 0.0089$, and its mean deformation is in fact slightly smaller. The tip-only ablation is worse than either, crushing the rim to $0.8272 \pm 0.0304$, a departure $11.3\times$ that of the PACT Hand and $1.38\times$ that of the LEAP Hand.

\looseness=-1 Torque dependence separates the three hands. Neither the PACT Hand, at $-0.0055$ per \si{N.m}, nor the LEAP Hand, at $-0.101$, deforms the cup measurably more as the cap rises, but the tip-only hand does, at $-0.467$. Contact concentrated on fingertips converts torque straight into deformation. Trial-to-trial spread shows a similar trend. The spread of the LEAP Hand and of the tip-only design is $6.1\times$ and $15\times$ that of the PACT Hand, although the same recorded grasp is replayed every time. Neither baseline, therefore, makes repeatable contact with the cup.
Overall, the PACT Hand held the compliant cup with the least deformation of the three hands, at a level comparable to the human hand. This supports the view that a hand optimized to imitate human grasping posture makes contact over a broader area, and that such a hand can handle a compliant object gently without tactile sensing.

\begin{figure}[t]
  \centering
  \NF{\includegraphics[height=3.4cm]{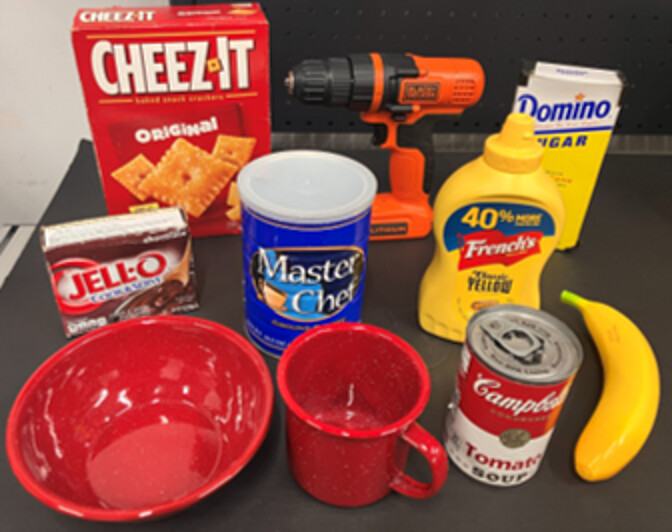}}\hfill
  \NF{\includegraphics[height=3.4cm]{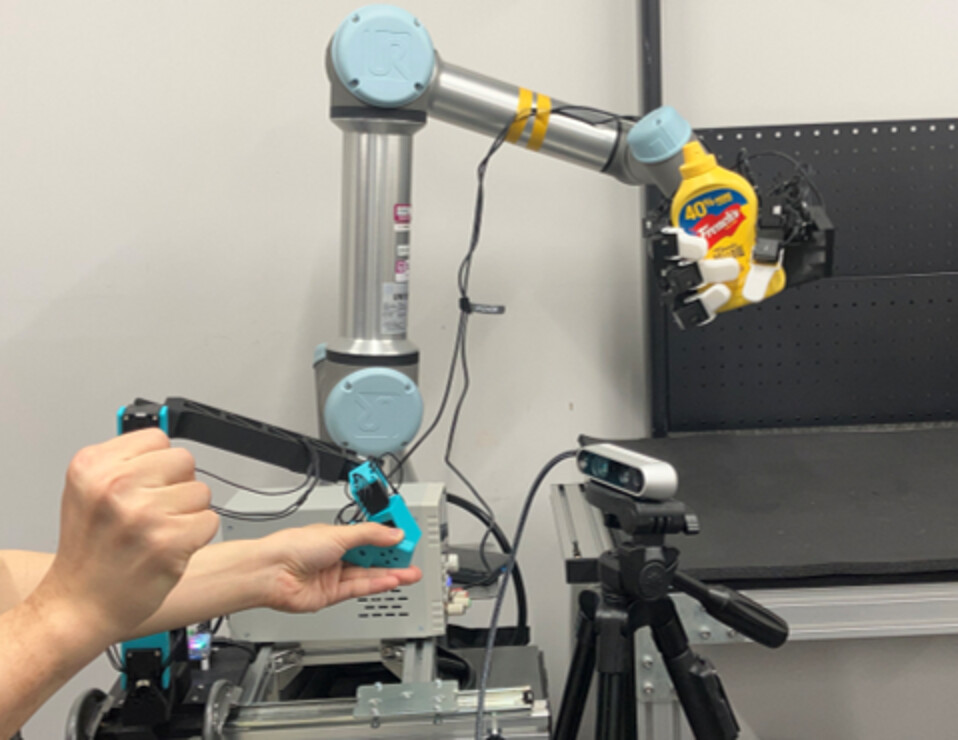}}
  \caption{Everyday-object comparison. Left: the ten YCB
  objects. Right: the teleoperation setup, operator with a
  GELLO leader arm, one RGB camera, and the UR5e-mounted hand.}
  \label{fig:ycbsetup}
\end{figure}

\begin{figure*}[t]
  \centering
  \NF{\includegraphics[width=1.0\textwidth]{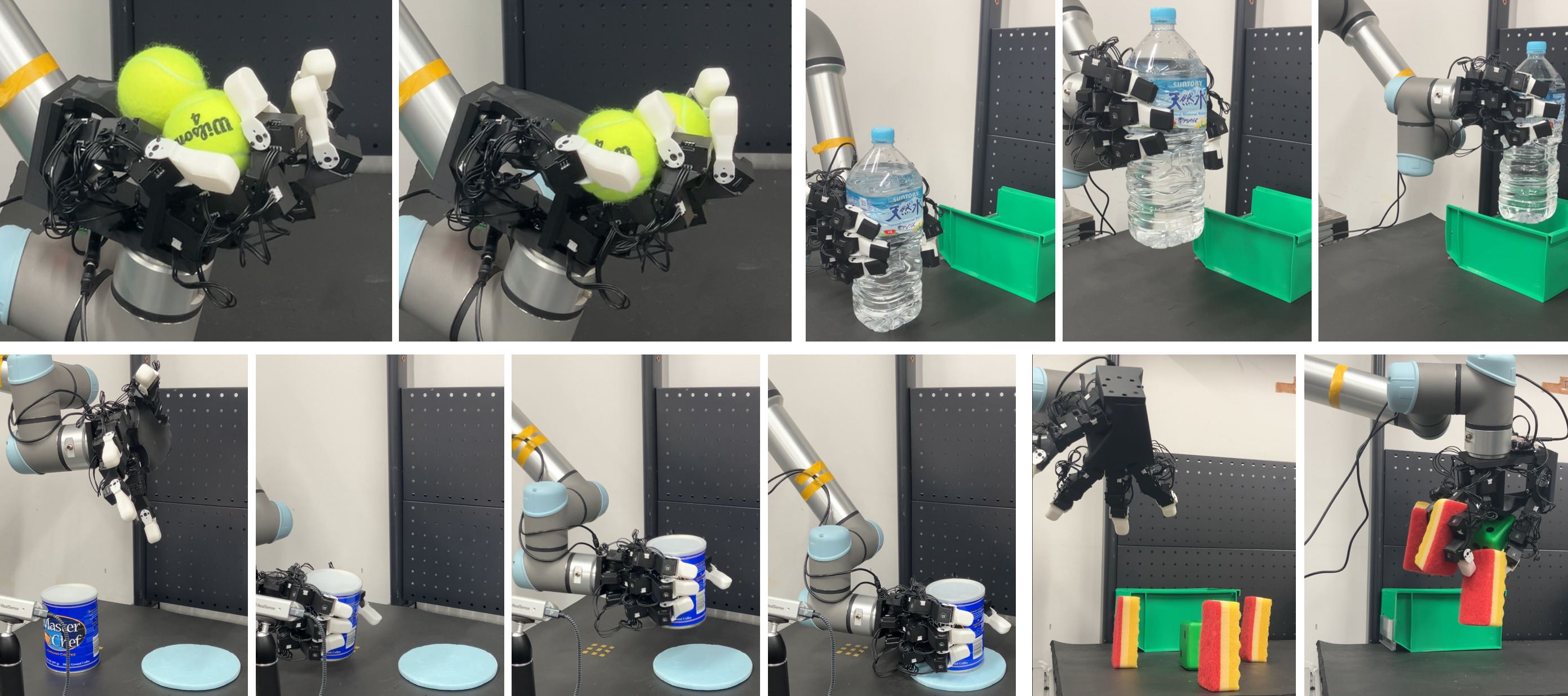}}
  \caption{Qualitative demonstrations. Top: in-hand stashing
  of four objects, and two tennis balls rolled on the palm.
  Bottom left: a \SI{2}{L} bottle grasped, lifted, and placed.
  Bottom right: the ACT policy autonomously picking the can and
  placing it on the target.}
  \label{fig:demos}
\end{figure*}

In the third experiment, three hands grasp ten physical YCB objects (Fig.~\ref{fig:ycbsetup}), selected in advance of all trials. The object set includes seven of the eight objects used in the design optimization, the scissors being unavailable, plus three objects never seen during optimization, namely the mug, mustard bottle, and power drill. Their weights spanned 74 to \SI{859}{g}. 
One grasp strategy per object, chosen in advance, is used by all three hands, since the objects differ in size and shape.
The cracker box was grasped across its shortest edge, because the LEAP Hand could not grasp it reliably across the longer edge. The banana and bowl carry grip tape, identical for every hand. The mug's handle is inaccessible at $1.56\times$ scale, so it is grasped from above. Each object was placed at the same position and orientation in each hand at the start of every trial. A single operator teleoperates live at a fixed \SI{0.167}{N.m} cap. A trial succeeds if the hand lifts the object and holds it for \SI{3}{s}. Each object and hand receives five scored trials.

\begin{table}[t]
\caption{Everyday-object grasping, successes out of five scored
trials at a cap of \SI{0.167}{N.m}. $\dagger$ marks objects unseen
during design. $\ddagger$ marks objects using grip tape.}
\label{tab:ycb}
\centering\footnotesize\renewcommand{\arraystretch}{0.93}
\setlength{\tabcolsep}{5pt}
\NF{\begin{tabular}{lccc}
\toprule
Object & PACT & LEAP & Tip-only \\
\midrule
master chef can            & \textbf{5/5} & 3/5 & 2/5 \\
cracker box                & \textbf{4/5} & 2/5 & 0/5 \\
sugar box                  & 5/5 & 5/5 & 4/5 \\
tomato soup can            & 2/5 & \textbf{4/5} & 0/5 \\
pudding box                & \textbf{5/5} & 4/5 & \textbf{5/5} \\
banana$^{\ddagger}$        & \textbf{5/5} & 3/5 & 0/5 \\
bowl$^{\ddagger}$          & \textbf{4/5} & 3/5 & \textbf{4/5} \\
mug$^{\dagger}$            & \textbf{3/5} & 1/5 & 0/5 \\
mustard bottle$^{\dagger}$ & \textbf{5/5} & 2/5 & 0/5 \\
power drill$^{\dagger}$    & 5/5 & 5/5 & 1/5 \\
\midrule
all & \textbf{43/50 (86\%)} & 32/50 (64\%) & 16/50 (32\%) \\
\bottomrule
\end{tabular}}
\end{table}

As a result, the PACT Hand succeeds in 43/50 trials (86\%), against 32/50 for the LEAP Hand (64\%) and 16/50 for the tip-only design (32\%), as listed in Table~\ref{tab:ycb}. Against the LEAP Hand, the PACT Hand leads on seven of the ten objects and trails on one. The exception is the slender soup can, where the PACT Hand's $1.56\times$ fingertips are oversized. Against the tip-only design it leads on eight objects and never trails, and the tip-only design falls below the LEAP Hand as well.

Success rate alone underestimates the performance gap, because a trial may also fail by exceeding the allotted time limit. The mean teleoperation time per successful trial, computed as the total teleoperation time divided by the number of successes, is \SI{10.6}{s} for the PACT Hand, \SI{21.8}{s} for the LEAP Hand, and \SI{43.5}{s} for the tip-only design. When unfinished trials are counted at the time limit, the median trial duration is \SI{8.1}{s} and \SI{12.8}{s} for the first two hands, but the full \SI{60}{s} for the tip-only design. More than half of the tip-only trials fail to achieve any grasp, and the PACT Hand is faster on all ten objects.

\looseness=-1 The observed failure modes further distinguish the three hand designs. LEAP Hand exhibits slipping, in-hand object twisting, and involuntary finger motions when its inverse kinematics solver jumps between different joint configurations that reach the same fingertip target. While LEAP Hand can initiate grasps, it fails to maintain stable holds.
The tip-only hand's dominant failure mode is object slippage between the fingers, occurring in four out of five soup-can trials and additionally on the mustard bottle, drill, and banana. This hand never forms a continuous holding surface. In contrast, the PACT Hand's failures are confined to the slender soup can and the top-down grasp of the mug. Since the heaviest object, at \SI{859}{g}, remains below the \SI{1}{kg} capacity measured for the LEAP Hand in the payload experiment above, no object tested here approaches the load-bearing limit of any hand. Consequently, the tip-only hand's 32\% success rate reflects a contact failure rather than a load failure, and the PACT Hand's residual weakness lies in small-feature precision at $1.56\times$ scale.

\looseness=-1 The fourth demonstration adds arm-hand teleoperation with a GELLO leader arm~\cite{gello2023}, shown in Fig.~\ref{fig:demos} and the supplementary video. Four everyday objects are stashed in the hand, a \SI{2}{L} bottle of approximately \SI{2.0}{kg} is lifted, and two tennis balls are rolled on the palm. As a demonstration of the intended use rather than as a contribution, an ACT policy~\cite{zhao2023act} trained on 54 teleoperated demonstrations of can-to-tray pick-and-place achieved a 94\% success rate (17/18 trials) at taught placements and 75\% (9/12 trials) at held-out ones.
\section{Conclusion}
\label{sec:conclusion}

We presented a co-design of the deployed retargeter and its mechanism, a hand tuned jointly with the posture-IK law that governs it. This single law serves at once as the design objective, the evaluation metric, and the runtime controller. The PACT Hand reproduces human posture to within \SI{3.9}{mm} of kinematic error, a third of the baseline's error, with no gap between training and test conditions. With equal per-joint torque limits, it carries twice the payload, deforms a paper cup seven times less (comparable to a human hand), and succeeds on 86\% of everyday-object grasps versus 64\%. A hand built from the baseline's own servos and optimized only for fingertip objectives reaches the same payload ceiling, yet deforms the cup more and grasps half as many objects. The objective thus sacrifices contact rather than force, and performance depends on the full-finger correspondence, not on the co-design apparatus alone.

For the twelve YCB objects outside the design set, tracking accuracy drops by only \SI{0.16}{mm}, showing that the geometry is not fitted to object identity. Yet the one grasp type absent from the design set, handle grasping, proved most costly, at \SI{4.74}{mm} on the pitcher. A missing grasp type is therefore a genuine gap, whereas a missing object is not. The test poses come from the same subjects on a shared mean hand shape, so robustness to operator hand size remains untested. The physical evaluation uses open-loop teleoperation, and the rule family is a linear seed refined via damped posture IK. A natural next step is to augment the objective with differential measures of the deployed map, capturing how it amplifies operator noise.

\bibliographystyle{IEEEtran}
\bibliography{refs}

\end{document}